\documentclass[10pt,twocolumn,letterpaper]{article}

\newcount\Comments  
\usepackage{color}
\definecolor{darkgreen}{rgb}{0,0.5,0}
\definecolor{purple}{rgb}{1,0,1}
\definecolor{rust}{rgb}{0.7,0.2,0.2}

\usepackage{iccv}
\usepackage{times}
\usepackage{epsfig}
\usepackage{graphicx}
\usepackage{amsmath}
\usepackage{amssymb}
\usepackage{multirow}
\usepackage{float}
\usepackage{listings}
\usepackage{enumitem}

\usepackage[breaklinks=true,bookmarks=false]{hyperref}

\iccvfinalcopy 

\def\iccvPaperID{94} 

\ificcvfinal\fi

\begin{document}


\title{Polygon Intersection-over-Union Loss for Viewpoint-Agnostic Monocular 3D Vehicle Detection}

\author{Derek Gloudemans\\
Vanderbilt University\\
{\tt\small derek.gloudemans@vanderbilt.edu}
\and
Xinxuan Lu\\
Vanderbilt University\\
{\tt\small xinxuan.lu@vanderbilt.edu}
\and
Shepard Xia\\
Vanderbilt University\\
{\tt\small weitao.xia@vanderbilt.edu}
\and
Daniel Work\\
Vanderbilt University\\
{\tt\small dan.work@vanderbilt.edu}
}

\maketitle
\ificcvfinal\thispagestyle{empty}\fi

\begin{abstract}
   Monocular 3D object detection is a challenging task because depth information is difficult to obtain from 2D images. A subset of \textit{viewpoint-agnostic} monocular 3D detection methods also do not explicitly leverage scene homography or geometry during training, meaning that a model trained thusly can detect objects in images from arbitrary viewpoints. Such works predict the projections of the 3D bounding boxes on the image plane to estimate the location of the 3D boxes, but these projections are not rectangular so the calculation of IoU between these projected polygons is not straightforward. This work proposes an efficient, fully differentiable algorithm for the calculation of IoU between two convex polygons, which can be utilized to compute the IoU between two 3D bounding box footprints viewed from an arbitrary angle. We test the performance of the proposed polygon IoU loss (PIoU loss) on three state-of-the-art viewpoint-agnostic 3D detection models. Experiments demonstrate that the proposed PIoU loss converges faster than L1 loss and that in 3D detection models, a combination of PIoU loss and L1 loss gives better results than L1 loss alone (+1.64\% $AP_{70}$ for MonoCon on cars, +0.18\% $AP_{70}$ for RTM3D on cars, and +0.83\%/+2.46\% $AP_{50}/AP_{25}$ for MonoRCNN on cyclists).  
\end{abstract}

\vspace{-0.15in}
\section{Introduction}
Autonomous driving is a primary domain that propels research in 3D object detection. Precise detection and localization of vehicles and pedestrians within a driving scenario are paramount to autonomous vehicles functioning safely and effectively. To enable this end, densely annotated ego-vehicle driving datasets produced with carefully calibrated and heavily instrumented test vehicles such as KITTI \cite{geiger_are_2012}, NuScenes \cite{caesar2020nuscenes}, Waymo OpenDrive \cite{sun2020scalability} have enabled a large  body of work on 3D object detection and tracking tasks. As a result, extremely accurate detection 3D vehicle, cyclist, and pedestrian models have been proposed that leverage the full suite of available sensors, including LIDAR, stereo images, and depth images \cite{lang_pointpillars_2019} \cite{maturana_voxnet_2015} \cite{qi_frustum_2018} \cite{shi_pointrcnn_2019} \cite{shi_pv-rcnn_2020} \cite{yan_second_2018}.  State-of-the-art 3D detection methods on KITTI frequently score above 90\% $AP_{70}$ (average precision) \cite{geiger_are_2012}.

The dense sensor set provided in these datasets comes at a price. Methods proposed utilizing these works are not generalizable to other vehicles with different or less capable sensing. Recognizing this shortcoming, \textit{monocular 3D object detection} methods have been proposed to predict object positions using a single camera and no additional sensors \cite{he_mono3d_2019, roddick2018orthographic,hu2019joint}. Posing the 3D object detection problem in this manner introduces the challenge of recovering depth information from an image, which is inherently depth-ambiguous. Monocular methods take a step in the direction of generality; only a single camera is required for detection; yet these methods incorporate information from the 3D scene explicitly (e.g., into the model anchor box generation architecture) or implicitly (by training to regress object positions directly in 3D space). Thus, a model trained for one camera in one vehicle can't be easily applied to another vehicle and camera, and training data is only available for a very small subset of instrumented vehicles.

\begin{figure*}[htb]
    \centering
    \includegraphics[width=\textwidth]{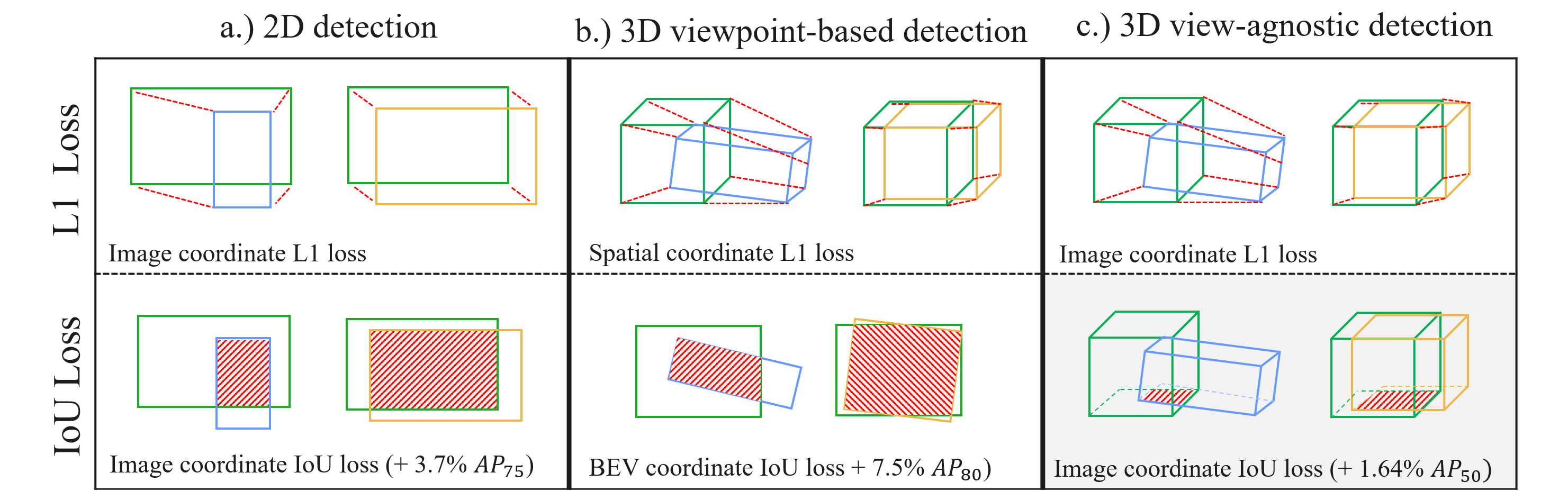}
    \caption{Example of IoU-based losses and relative improvements versus L1-loss in a.) image coordinates, where bounding boxes are axis-aligned, \cite{rezatofighi_generalized_2019} b.) 3D viewpoint-based detection, in which the scene homography can be used to precisely compute object rectangular footprints in a \textit{bird's-eye view} \cite{zhou_iou_2019} and c.) 3D viewpoint-agnostic detection (this work), using 3D box projections into 2D space.}
    \label{fig:IoU-methods}
    \vspace{-0.15in}
\end{figure*} 

To detect 3D objects from a single, arbitrary camera, such as a cell-phone camera or dashcam from an arbitrary vehicle, a more general method is required. Recently, a subset of monocular 3D detection methods have attempted to detect 3D bounding boxes for vehicles without utilizing scene information in the trained model. Instead, these models predict positions natively in 2D image space and incorporate scene \textit{homography} only after training and after inference \cite{liu_learning_2022,sochor2016boxcars,li_rtm3d_2020,shi_geometry-based_2021}. By posing 3D detection in this way, a trained model may be able to generalize to an unseen camera view simply by changing the post-inference scene homography. In other words, these methods are \textit{ viewpoint-agnostic}. The performance of these models, while steadily improving, still trails other monocular 3D detection methods and 3D detection methods more generally.

This  work seeks to leverage a key trend in object detection works: across a variety of domains and detection problem formulations, \textit{intersection-over-union} (IoU) based methods have been shown to outperform L1 and L2 norm-based methods for loss calculation during training. Intersection over Union (IoU) is commonly used for measuring the similarity between (generally) two rectangular, axis-aligned bounding boxes. Many previous works on 2D object detection tasks demonstrate that IoU can be used as a regression loss for axis-aligned 3D bounding boxes. In \cite{rezatofighi_generalized_2019} \cite{zheng_distance-iou_2020} and \cite{zheng_enhancing_2022} it is shown that incorporating IoU loss into 2D object detection models can improve their performance. In a similar vein, \cite{zhou_iou_2019} and \cite{zheng2020rotation} show that IoU for two rotated rectangles can be used as a loss function to improve the performance of 3D object detection models. Figure~\ref{fig:IoU-methods} summarizes. Unfortunately, these methods are not directly applicable to viewpoint-agnostic monocular methods because the projection of a 3D bounding box into an image does not result in rectangular planes; rather, the six surrounding planes of vehicles occupy arbitrary quadrilaterals in pixel-space. Thus, most existing methods use L1 loss to regress the eight corner points of the 3D box on 2D image planes.

The core contribution of this work is to present a new and efficient way of calculating the IoU between two convex polygons which we refer to as \textit{polygon IoU} and implement it as a loss function (PIoU loss).  We show both in simulation and in 3 state-of-the-art viewpoint-agnostic 3D detection models that the loss function converges faster than L1 loss. We implement a batched version of the IoU function between two polygons to enable fast training of models with the method. We utilize models trained with PIoU loss on the KITTI 3D detection benchmark and show that, in most cases, the new loss formulation increases model accuracy, particularly for higher requisite IoU thresholds. 

The rest of this paper is laid out as follows: Section \ref{related work} reviews related works, Section \ref{polygon iou loss} describes the PIoU method in more detail. Section~\ref{experiments simulated} describes experiments comparing L1 loss and PIoU loss with simulated polygons, Section \ref{experiments KITTI} details experiments on the KITTI benchmark and describes implementation details for incorporating  PIoU loss into 3 detection models. Section \ref{sec:results} describes the results.

\section{Related Work} \label{related work}


\subsection{Monocular 3D object detection}
Monocular 3D detection methods seek to generate a set of 3D bounding boxes in 3D space based on a single camera image. One early work is Mono3D \cite{chen_monocular_2016}, which generates rich 3D proposals with the assumption that vehicles locate on the ground plane and then scores the boxes with contextual information and size, location, and shape priors. Likewise, \cite{roddick2018orthographic} generates 3D proposals and ensures that feature map computations are orthographic such that objects further away and occupying fewer pixels do not occupy less of the final feature map space. In \cite{hu2019joint}, detection and object tracking are accomplished by directly regressing 3D coordinates, but anchor boxes are generated in 2D image-space (thus the scene homography is implicitly learned during training).

\subsection{Viewpoint-Agnostic Monocular 3D Detection}
Viewpoint agnostic monocular methods can roughly be divided into two categories: i.) methods that regress 2D bounding boxes or segmentations along with  augmenting outputs and utilize homography constraints to subsequently predict 3D outputs, and ii.) methods that regress 3D projections of keypoints or bounding box corner points.

In the first category, Deep3DBox \cite{mousavian_3d_2017} predicts a 2D bounding box, the observation angle, 3D object size, and object 3D center position (in the image) from the features enclosed by the 2D bounding boxes, as the bounding box can subsequently be fit by the constraint that its 2D projection falls within the 2D bounding box. Shift R-CNN \cite{naiden_shift_2019} and Cascade Geometric Constraint \cite{fang_3d_2019} leverage the fact that 4 vertices of the 3D bounding box must lie on the 2D bounding box. The main drawback of these models is that they rely on accurate predictions of 2D bounding boxes. Errors in 2D bounding boxes compound in the 3D prediction. Likewise, \cite{chabot2017deep} utilizes a 2D bounding box and manipulates a simplified 3D vehicle model to optimize the 3D object position within the bounding box. 3D-RCNN \cite{kundu_3d-rcnn_2018} takes additional segmentation inputs and generates a compact 3D representation of the scenes. It exploits class-specific shape priors by learning a low-dimensional shape-space from collections of CAD models. 

Most recent methods fall in the latter category, representing vehicles as polyhedrons or 3D bounding boxes.  Mono3D++ \cite{he_mono3d_2019} represents a vehicle as 14 keypoints and learns the 2D keypoints using EM-Gaussian method. MonoRCNN \cite{shi_geometry-based_2021} is built upon Faster R-CNN and adds 3D attribute and distance heads to recover 3D bounding boxes. The heatmap concepts proposed by CenterNet \cite{zhou_objects_2019} inspired many monocular 3D detection models because this model's structure is well-suited to keypoint regression. RTM3D \cite{li_rtm3d_2020} uses CenterNet-based structures to regress nine keypoints of the 3D bounding box corresponding to the eight corners and the center of the 3D cuboids. RTM3D also regresses distance, 3D box dimension, and orientation of vehicles, then solves an optimization for the best-fitting bounding box in 3D space for each object. Likewise, Monocon \cite{liu_learning_2022} and Monoflex \cite{zhang_objects_2021} are built upon CenterNet. Monoflex directly predicts 3D dimensions and orientations and incorporates multiple depth estimation methods to increase accuracy.  In \cite{sochor2016boxcars}, no scene information is ever used, and instead the vanishing points and, thus, scene homography are directly computed from output 3D bounding boxes (albeit in a traffic monitoring context). 

\subsection{IoU Loss in Object Detection}
L1 and L2 losses are widely used in object detection models but ignore the shape of bounding boxes and are easily influenced by the scales of boxes. Conversely, IoU encodes the shape properties of objects and is invariant to the scale. Thus, IoU-based loss formulations have achieved good performance in object detection. In \cite{rezatofighi_generalized_2019}, Generalized Intersection over Union (GIoU) loss is proposed to provide better convergence for non-overlapping bounding boxes. The authors incorporate GIoU loss into YOLO v3 \cite{redmon_yolov3_2018}, Faster R-CNN \cite{ren_faster_2015}, and Mask R-CNN \cite{he_mask_2017} and show a consistent improvement in their performance on popular object detection benchmarks such as PASCAL VOC and MS COCO. \cite{zheng_distance-iou_2020} similarly incorporates the distance between bounding boxes to aid convergence in non-overlapping cases. In \cite{zheng_enhancing_2022}, the authors introduce Complete-IoU (CIoU) loss to consider three geometric factors: overlap area, normalized central point distance, and aspect ratio. CIoU loss is used in YOLOv4 \cite{bochkovskiy_yolov4_2020} and leads to notable gains of average precision (AP) and average recall (AR). In \cite{zhou_iou_2019} and \cite{zheng2020rotation}, IoU loss is defined for two rotated bounding boxes into several 3D object detection frameworks. which leads to consistent improvements for both bird-eye-view 2D detection and point cloud 3D detection on the public KITTI benchmark \cite{geiger_are_2012}.

While promising, these IoU loss variants are not suitable for the keypoints of 3D bounding boxes projected to the image plane. No work yet analyzes the performance of incorporating IoU loss into viewpoint-agnostic monocular 3D detection frameworks.

\section{Polygon IoU Loss} \label{polygon iou loss}
 The \textit{Polygon IoU} (PIoU) method proposed calculates the intersection-over-union metric for any two convex polygons in 2D coordinates. The inputs are two sets $\mathcal{A}$ and $\mathcal{B}$ consisting of the $(x,y)$ corner coordinates of each polygon, and the algorithm output falls in the range $[0,1]$. This output can then be utilized as a loss function $ Loss = 1 - PIoU$.

\subsection{Overview}
 Polygon IoU loss calculation consists of:
\begin{enumerate}[topsep=0pt,itemsep=-1ex,partopsep=1ex,parsep=1ex]
    \item Order the points of $\mathcal{A}$ and $\mathcal{B}$ clockwise.
    \item Compute $\mathcal{C}$, the set of all points of intersection of any two edges of the polygons.
    \item Find $\mathcal{A}_B$, the set of  all points in $\mathcal{A}$ that lie in the interior of $\mathcal{B}$,  and vice versa for $\mathcal{B}_A$.
    \item Compute the area of the convex polygon defined by the overlapping set of points $\mathcal{I} = \mathcal{A}_B \cup \mathcal{B}_A \cup \mathcal{C}$
    \item Compute the areas of $\mathcal{A}$, $\mathcal{B}$ and $\mathcal{I}$.
    \item Compute PIoU according to:
    \vspace{-0.1in}
     \begin{equation} \label{eq:iou}
        \frac{Area_\mathcal{I}}{Area_\mathcal{A} + Area_\mathcal{B} - Area_\mathcal{I}}
     \end{equation}
     \vspace{-0.2in}

\end{enumerate}
We describe each step in more detail below.

\subsection{Clockwise}
A set of points is ordered in a clockwise manner by computing the geometric center of the polygon. Then, angles are calculated between an arbitrary first point (defined to be at 0\textdegree), and each other point, relative to the geometric center. Points are then sorted in order of decreasing angle relative to the center. Note that the clockwise ordering of $\mathcal{A}$ and $\mathcal{B}$ is necessary for subsequent computational steps which assume a clockwise, geometrically adjacent ordering of points.

\subsection{Finding intersections $\mathcal{C}$}
For a line that passes through points 
$(x_1, y_1), (x_2, y_2)$, and a line that passes through points  $(x_3, y_3), (x_4, y_4)$, the intersections $(I_x,I_y)$ are calculated as:
\begin{equation} \label{eq:I}
    \begin{aligned}
    I_x &= \frac{(x_1y_2-y_1x_2)(x_3-x_4)-(x_3y_4-y_3x_4)(x_1-x_2)}{D}, \\
    I_y &= \frac{(x_1y_2-y_1x_2)(y_3-y_4)-(x_3y_4-y_3x_4)(y_1-y_2)}{D}
    \end{aligned}
\end{equation}
\begin{equation} \label{eq:D}
    D=(x_1-x_2)(y_3-y_4)-(y_1-y_2)(x_3-x_4)
\end{equation}
Utilizing this formula, the intersection point between every line defined by consecutive points in $\mathcal{A}$ and  $\mathcal{B}$ is computed. Some of these intersections do not lie on the polygons $\mathcal{A}$ and  $\mathcal{B}$. We filter invalid points and only keep the intersections with x coordinates within the range of both two pairs of points in $\mathcal{C}$. x must satisfy $x_1 \leq I_x \leq x_2$ and $x_3 \leq I_x \leq x_4$. (Restrictions on y coordinates are automatically satisfied if x coordinates are within the correct range.)

\subsection{Finding points $\mathcal{A}$ inside $\mathcal{B}$} 
\label{inside}
\begin{figure}
    \centering
    \includegraphics[width=50mm]{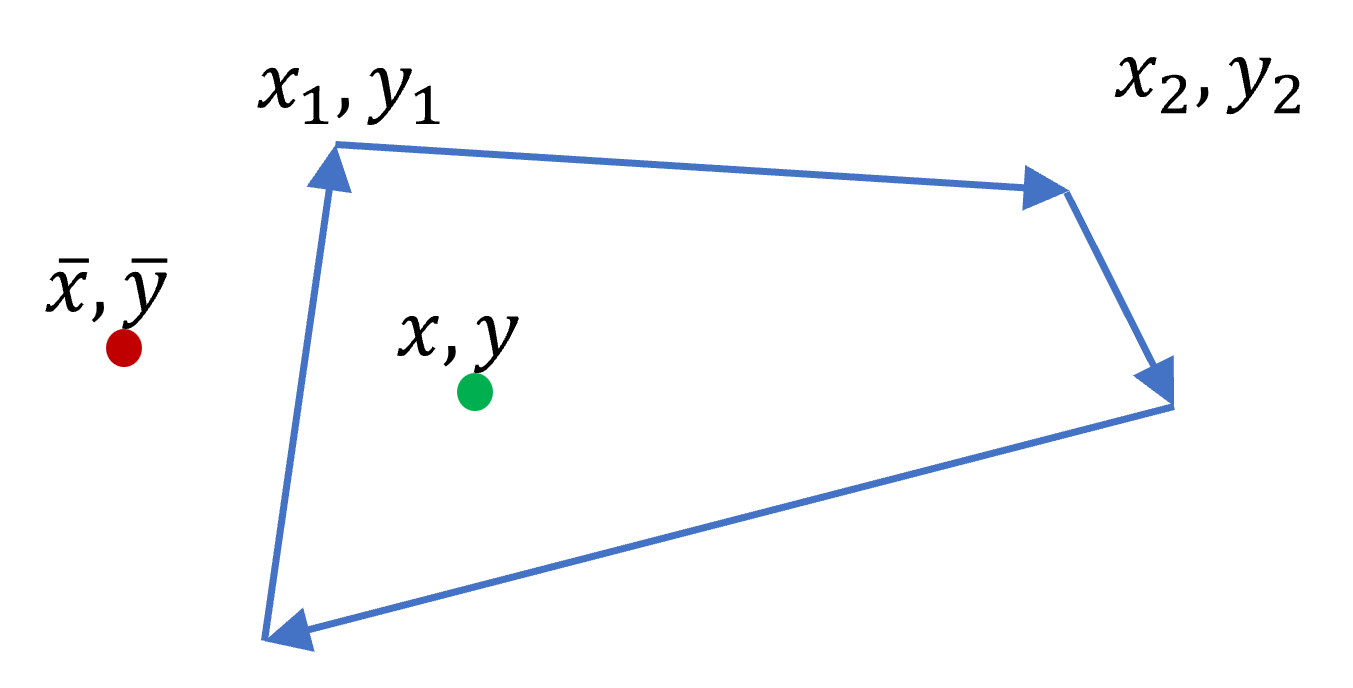}
    \caption{A polygon with edges marked with directions in clockwise order. Relative to each line segment, the point $(x,y)$ lies on the right side, while $(\bar{x},\bar{y})$ lies on the left of one line segment and on the right of three line segments.
    }
    \label{fig:polygon}
    \vspace{-.15in}
\end{figure}

For a convex polygon, each edge is assigned a direction in clockwise order, as in Figure \ref{fig:polygon}. Then if and only if a point $(x,y)$ lies on the same side of all the edges of the polygon, it lies inside the polygon. Let the endpoints of the line segment be $(x_1, y_1)$ and $(x_2, y_2)$.  We compute:
\begin{equation}
    (y - y_1) (x_2 -x_1) - (x - x_1) (y_2 - y_1)
\end{equation}
where a positive result means that the point lies on the left of the line, a negative result means that the point lies on the right of the line, and zero means that the point lies on the line. Each point in $\mathcal{A}$ is checked against the line segments defined by $\mathcal{B}$ to determine $\mathcal{A}_B$, the set of points in $\mathcal{A}$ lie within $\mathcal{B}$, and the opposite is done to determine $\mathcal{B}_A$. The full set of points defining the intersection of $\mathcal{A}$ and $\mathcal{B}$ is then defined by  $\mathcal{I} = \mathcal{A}_B \cup \mathcal{B}_A \cup \mathcal{C}$.

\subsection{Calculating the area of a polygon} \label{calculate area}
Let $(x_i, y_i)$ represent the coordinates of the i-th point of a polygon. Let
\begin{equation}
    \begin{aligned}
        \boldsymbol{x}=[x_1, x_2, ..., x_n]^T&, \boldsymbol{x^*}=[x_2, x_3, ..., x_n, x_1]^T, \\
        \boldsymbol{y}=[y_1, y_2, ..., y_n]^T&, \boldsymbol{y^*}=[y_2, y_3, ..., y_n, y_1]^T
    \end{aligned}
\end{equation}
The area of a polygon is computed by:
\begin{equation} \label{eq:area}
    Area=\frac{1}{2}(\boldsymbol{x}^T\boldsymbol{y^*} - \boldsymbol{y}^T\boldsymbol{x^*})
\end{equation}

The areas of  $\mathcal{I}$,  $\mathcal{A}$, and $\mathcal{B}$ are thus computed, and equation \ref{eq:iou} is used to determine the PIoU.

\subsection{Batched implementation}
Polygon IoU loss is applicable for convex polygons with any number of corners. However, a variable number of points among polygons impedes calculation in a batched, vectorized implementation. Thus, for batched implementation, we restrict inputs to a fixed number of points per batch (in practice, for 3D bounding box calculation, all polygons will be four-sided). For 4-sided polygons, there are at most 8 points in the set  $\mathcal{C}$, at most 4 interior points in each set $\mathcal{A}_B$ and  $\mathcal{B}_A$, and at most 8 points in $\mathcal{I}$. So, the size of the vector that represents the intersection region is set to 8. If the actual number of points in the set  $\mathcal{I}$ is less than 8, the set is padded with repeated points which won't alter the result of~\eqref{eq:area}. 

For polygons with $P$ points, the maximum number of corner points in $\mathcal{I}$ is set to be $2P$. The batched implementation of a function empirically computes forward and backward training passes significantly faster than a non-batched loop-based implementation of PIoU. Pseudo-code for a batched, vectorized implementation of PIoU with batch size $B$ is given below. The shape of the output tensor at each step is given in square brackets.

\noindent\rule{8cm}{0.4pt}
\textbf{Algorithm: Batched Tensor PIoU} \newline
Inputs: $polygon A$ and $polygon B$. Each [$B$,$P$,2]. \newline
$\triangleright$    $\mathcal{C}$ = intersections of all line segments in $polygon A$ and $polygon B$ using (0, 0) to fill the empty. [$B$,$2P$,2]. \newline
$\triangleright$    $\mathcal{A}_B$ = all points of $polygon A$ that are inside $polygon B$, using (0, 0) to fill the empties. [$B$,$P$,2].\newline
$\triangleright$    $\mathcal{B}_A$ = all points of $polygon B$ that are inside $polygon A$ using (0, 0) to fill the empties. Tensor shape: [$B$,$P$,2]. \newline
$\triangleright$    $overlap$ = union of [$\mathcal{C}$, $\mathcal{A}_B$, $\mathcal{B}_A$]. [$B$,$4P$,2]. \newline
$\triangleright$    sort $overlap$ in decreasing order with respect to the distance from (0, 0) \newline
$\triangleright$    keep the first 8 points in $overlap$. [$B$,$2P$,2].\newline
$\triangleright$    $placeholder$ = the points farthest from (0, 0) in $overlap$. [$B$,1,2].\newline
$\triangleright$    replace (0, 0) in $overlap$ with $placeholder$ \newline
$\triangleright$  $areaO$,$areaA$,$areaB$ = areas of $overlap$, $polygon A$, and $polygon B$. Each [$B$]. \newline
$\triangleright$    PIoU = $areaO / (areaA+areaB-areaO)$. [$B$]. \newline
\noindent\rule{8cm}{0.4pt}

\begin{figure*}[ht]
    \centering
    \includegraphics[width=170mm]{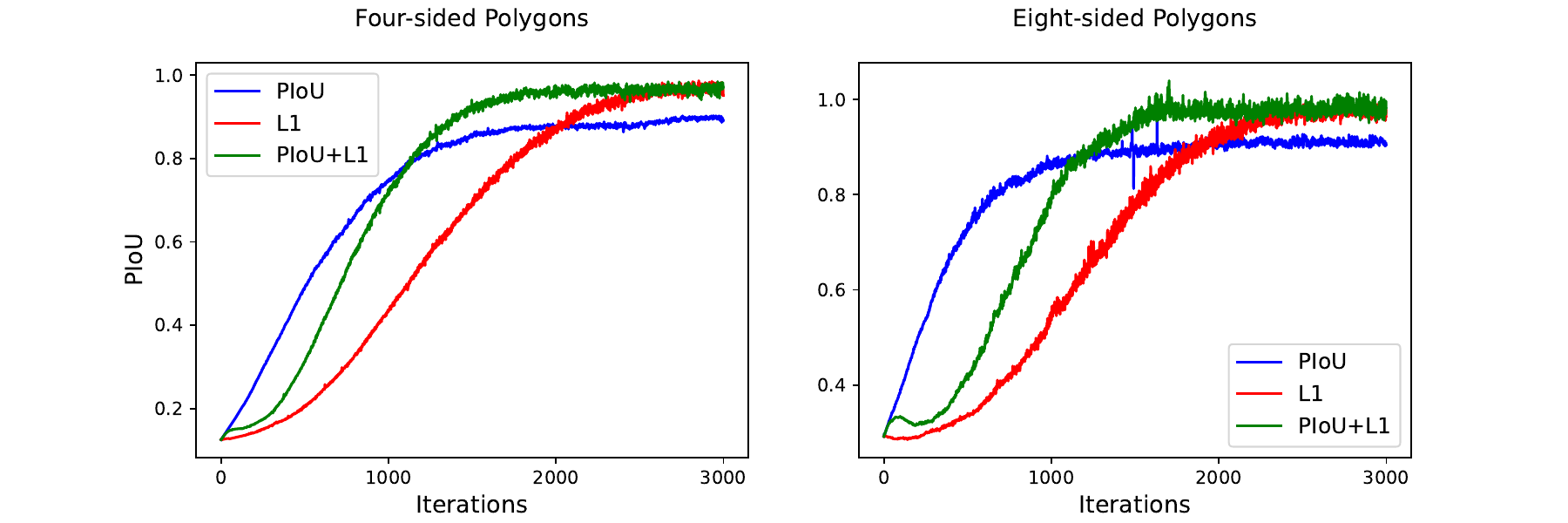}
    \caption{PIoU score versus iteration for (left) 4-sided simulated polygons and (right) 8-sided unrestricted simulated polygons.}
    \label{fig:fig1}
\end{figure*}

\vspace{-0.05in}

\subsection{Edge Cases}
When an edge from polygon A and an edge from polygon B are parallel, the intersections between the two edges are ill-defined because the denominator in~\eqref{eq:I} approaches zero. This occurs when: i.) two edges coincide with the same line. ii.) Two edges are parallel but not coincident with the same line. In case i.), the points of $\mathcal{A}$ and $\mathcal{B}$ already suitably define the intersection points, so including points of $\mathcal{A}$ exactly on an edge of $\mathcal{B}$ in $\mathcal{A}_B$ covers this case. In case ii.), the two edges have no intersection. Thus, we can remove these intersections from $\mathcal{C}$ for numerical stability. 

\section{Experiments on simulated polygons} \label{experiments simulated}

\subsection{Four-sided convex polygons}
We generate two sets of quadrilaterals (4-sided polygons), with one set as the initial polygons and one set as the ground truth. The polygons are generated as centers and offsets of four points to ensure they are convex. We use the Adam optimizer to regress predicted polygons with the goal of approximating the ground truth polygons. The polygons are generated in a batch of 32 for the training. We compare the result of L1 loss, PIoU loss, and a combination of L1 and PIoU loss. The IoU with respect to iterations is plotted on the left of  Figure \ref{fig:fig1}. The results take the average of 5 independent trials. The PIoU loss converges the fastest in the beginning. However, the PIoU loss does not achieve a high IoU when it converges.  PIoU+L1 loss has the fastest convergence speed and accuracy after around 2000 iterations.

\subsection{Eight-sided unrestricted polygons}
We repeat this experiment with 8-sided polygons, this time not restricting the predicted polygons to be convex. IoU versus optimization iteration is plotted on the right of Figure \ref{fig:fig1}. The results take the average of 5 independent trials. The results are similar to the 4-sided polygon case. The PIoU loss converges the fastest initially, while PIoU+L1 loss converges faster than L1 loss alone and additionally reaches the highest overall IoU score. Non-convex polygons produce slightly more noise in loss curves, visible in the PIOU and PIOU+L1 curves of Figure \ref{fig:fig1} (right). This experiment on simulated 8-sided polygons shows that our PIoU loss has good performance even when the polygons are not convex and have more than four sides.

\subsection{Computation speed}
We compare the computation speed of our batched implementation of PIoU loss relative to a pixel-wise IoU loss (as used for object segmentation tasks). For a batch size of 1, our PIoU loss is 4.0x faster than the pixel-wise implementation. PIoU loss is 56.0x faster for a batch size of 16 and 281.6x faster for a batch size of 128.


\begin{figure*}[ht]
    \centering
    \includegraphics[width=2\columnwidth]{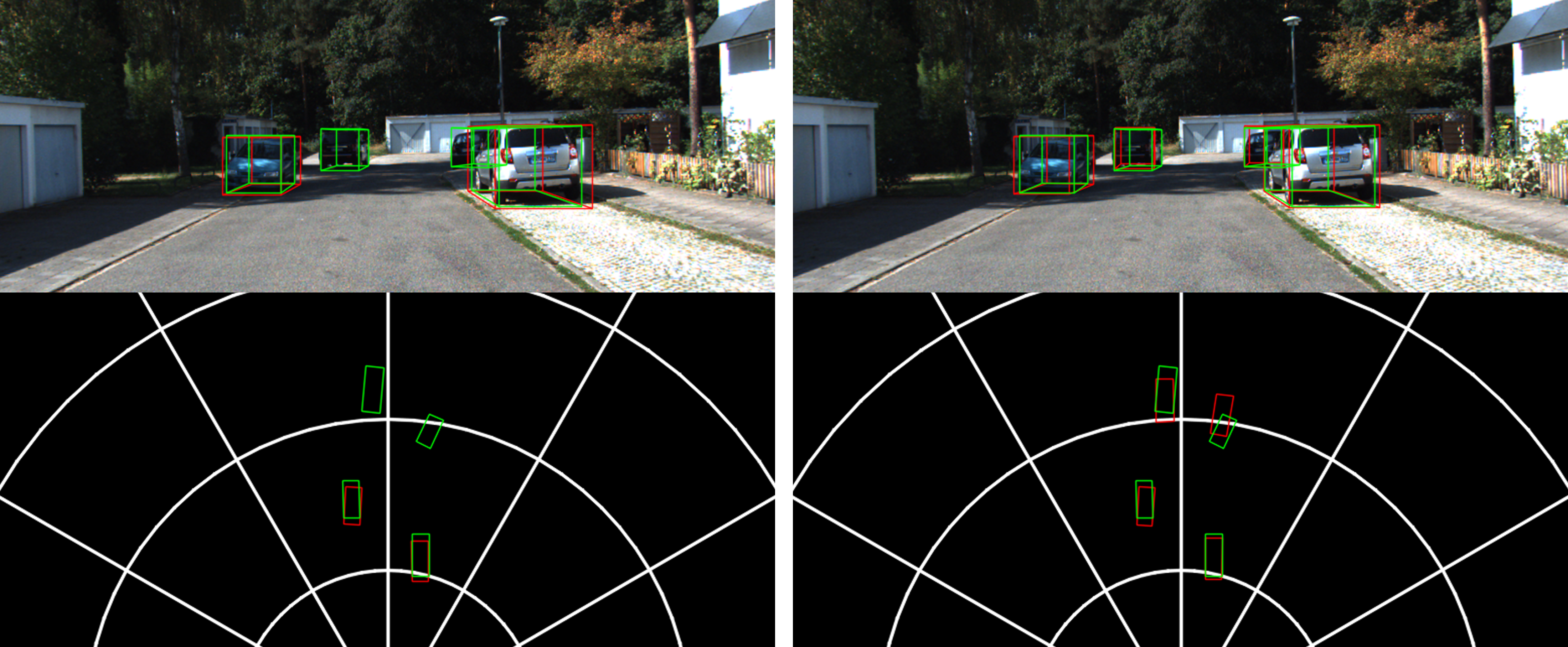}
    \caption{Predicted (red) and ground truth (green) 3D bounding boxes from the train/val split of KITTI dataset from MonoCon object detection model trained with L1 loss (left) and PIoU+L1 loss (right). Top images show the predicted bounding boxes in 3D space, and bottom images show the corresponding predicted footprints in a birds-eye view.} 
    \label{fig:visualizations}
    \vspace{-0.1in}
\end{figure*}

\section{Experiments on KITTI 3D} \label{experiments KITTI}
PIoU loss supports polygons with any number of points. However, there are not existing detection problems well suited to evaluating polygons with more than 4 points. We test it on 3D detection problems where the projection of 3D bounding boxes to the image plane can be separated into two quadrilaterals, the front 4 and back 4 corner coordinates of the 3D bounding box. We incorporate PIoU loss into RTM3D \cite{li_rtm3d_2020}, MonoCon \cite{liu_learning_2022}, and MonoRCNN \cite{shi_geometry-based_2021} and test each on the KITTI 3D benchmark \cite{geiger_are_2012}. In all three models, we compare the performance of using PIoU+L1 loss and only using L1 loss. The rest of this section describes the models, modifications, and experimental parameters for each experiment. The results are listed in Section \ref{sec:results}.

\subsection{Dataset}
We use KITTI 3D object detection benchmark as our training and evaluation dataset. As KITTI does not allow more than three submissions and the labels for the testing set are not released, we follow \cite{liu_learning_2022} to divide the official training set of KITTI 3D into 3712 training images and 3769 evaluation images. (We use \textit{train/val} to represent this split.) Three classes of objects, cars, pedestrians, and cyclists, are used for training. During training, only left-camera images and ground-truth labels are used (calibration matrices are used only to produce pixel-space 3D keypoint projections.) 
\subsection{Evaluation Metrics}
The objects in KITTI are categorized into easy, moderate, and hard according to their height, truncation level, and occlusion ratio. We evaluate the results for each category using the same evaluation guidelines as KITTI. As suggested by KITTI in 2019, we use 40 recall positions to calculate the average precision (AP) of results. KITTI sets different 3D bounding box overlap thresholds for cars (70\%) and cyclists (50\%). We evaluate cars at 70\%, 50\% and 30\% $AP_{3D}$ and cyclists at 50\% and 25\% $AP_{3D}$.

\subsection{Ensuring convexity}
As our polygon IoU loss is accurate when the two polygons are convex, we encourage the predicted keypoints to form convex polygons during training. The ground truth keypoints are projected from 3D cuboids, so it is guaranteed that they form two convex quadrilaterals. We  ensure that the initial predictions of the model are convex by adding a small offset (4 corner points of a square centered at the origin) to the predicted keypoints at initialization. During our training and testing, we find that a convex initial prediction is sufficient to make the PIoU loss converge.

\begin{table*}[ht]    
\begin{center}
\begin{tabular}{|l|lll|lll|lll|}
\hline
\multirow{2}{*}{Model + Loss Types} &
  \multicolumn{3}{l|}{$AP_{70}$} &
  \multicolumn{3}{l|}{$AP_{50}$} &
  \multicolumn{3}{l|}{$AP_{30}$} \\ \cline{2-10} 
 &
  \multicolumn{1}{l|}{Easy} &
  \multicolumn{1}{l|}{Mod} &
  Hard &
  \multicolumn{1}{l|}{Easy} &
  \multicolumn{1}{l|}{Mod} &
  Hard &
  \multicolumn{1}{l|}{Easy} &
  \multicolumn{1}{l|}{Mod} &
  Hard \\ \hline \hline \hline
RTM3D (L1) &
  \multicolumn{1}{l|}{3.88} &
  \multicolumn{1}{l|}{2.55} &
  1.96 &
  \multicolumn{1}{l|}{24.13} &
  \multicolumn{1}{l|}{\textbf{18.62}} &
  \textbf{15.34} &
  \multicolumn{1}{l|}{\textbf{51.31}} &
  \multicolumn{1}{l|}{40.04} &
  \textbf{35.03} \\ \hline
RTM3D (L1+PIoU) &
  \multicolumn{1}{l|}{\textbf{4.08}} &
  \multicolumn{1}{l|}{\textbf{2.70}} &
  \textbf{2.15} &
  \multicolumn{1}{l|}{\textbf{25.02}} &
  \multicolumn{1}{l|}{18.58} &
  15.25 &
  \multicolumn{1}{l|}{51.10} &
  \multicolumn{1}{l|}{\textbf{41.05}} &
  34.95 \\ \hline \hline \hline

MonoCon (L1) &
  \multicolumn{1}{l|}{22.27} &
  \multicolumn{1}{l|}{16.98} &
  14.69 &
  \multicolumn{1}{l|}{62.00} &
  \multicolumn{1}{l|}{46.19} &
  \textbf{41.71} &
  \multicolumn{1}{l|}{85.09} &
  \multicolumn{1}{l|}{65.94} &
  60.97 \\ \hline
MonoCon (L1+PIoU) &
  \multicolumn{1}{l|}{\textbf{24.93}} &
  \multicolumn{1}{l|}{\textbf{18.45}} &
  \textbf{15.49} &
  \multicolumn{1}{l|}{\textbf{63.88}} &
  \multicolumn{1}{l|}{\textbf{46.92}} &
  41.01 &
  \multicolumn{1}{l|}{\textbf{85.88}} &
  \multicolumn{1}{l|}{\textbf{66.60}} &
  \textbf{61.43} \\ \hline \hline \hline

  MonoRCNN (L1) &
  \multicolumn{1}{l|}{\textbf{16.94}} &
  \multicolumn{1}{l|}{\textbf{13.73}} &
  \textbf{11.67} &
  \multicolumn{1}{l|}{49.87} &
  \multicolumn{1}{l|}{\textbf{38.65}} &
  33.47 &
  \multicolumn{1}{l|}{\textbf{76.57}} &
  \multicolumn{1}{l|}{\textbf{60.93}} &
  \textbf{52.63} \\ \hline
MonoRCNN (L1+PIoU) &
  \multicolumn{1}{l|}{16.73} &
  \multicolumn{1}{l|}{13.62} &
  11.63 &
  \multicolumn{1}{l|}{\textbf{51.14}} &
  \multicolumn{1}{l|}{38.63} &
  \textbf{33.50} &
  \multicolumn{1}{l|}{74.05} &
  \multicolumn{1}{l|}{59.99} &
  51.85 \\ \hline
  
\end{tabular}
\end{center}
\caption{Results for RTM3D, MonoCon, and MonoRCNN on KITTI 3D Car on train/val split, evaluated by $AP_{3D}$ with IoU thresholds of 0.7, 0.5, and 0.3. Best result for each model at each threshold is shown in bold.}
\label{tab:cars-all}
\vspace{-0.15in}
\end{table*} 

\subsection{RTM3D}

\subsubsection{Experiment Settings}
We modify an unofficial implementation \cite{RTM3D-PyTorch} of RTM3D to do our experiments. We do not follow RTM3D to assume a Gaussian kernel for keypoints. We solve a least-squares problem to obtain the best-fitting 3D bounding boxes from the predicted keypoints, 3D dimension, and orientation. For comparison, We add our PIoU loss to regress the 2D projected keypoints of 3D bounding boxes. We use the same training settings for both the baseline and modified model. We use an Adam optimizer with an initial learning rate of 0.0002 which decreases by a factor of 0.1 in epochs 150 and 180. The weight decay is 1e-6. We train for a total of 200 epochs with a batch size of 16. The experiments run on Ubuntu 20.04 and RTX A5000. It took 14.3 hours to train the model with PIoU loss and 11.5 hours to train the model with L1 loss. Adding PIoU loss computation to the RTM3D model only slightly increases the training time.

\subsubsection{The Least-Squares Problem}
We define the least-squares problem for finding the best-fitting 3D bounding boxes similar to the definitions in RTM3D. For each predicted object, $\hat{P}$ represents the eight corner points of the 3D bounding box on the 2D image plane. $\hat{\Theta}, \hat{D}=[\hat{h}, \hat{w}, \hat{l}]^T, \hat{d}$ represent the predicted orientation, dimensions, and depth of the 3D bounding box. $T=[x,z,d]^T$ represents the position of the 3D bounding box, where $d$ represents the horizontal depth. $f(D, T, \Theta)$ maps the 3D bounding box to the 8 corner points in the image plane. We set $\alpha_P=0.05, \alpha_T=1, \alpha_D=1, \alpha_d=1$ in our experiments. The least-squares problem is defined as 
\begin{equation}
    \begin{aligned}
        \max\limits_{D, T, \Theta}&\alpha_P \|f(D, T, \Theta) - \hat{P}\|^2 + \alpha_T \|\Theta - \hat{\Theta}\|^2 \\
        + &\alpha_D \|D - \hat{D}\|^2 + \alpha_d \|d - \hat{d}\|^2
    \end{aligned}
\end{equation} 

\subsection{MonoCon}


\subsubsection{Experiment Settings}
We use an unofficial implementation \cite{monocon_unofficial} of MonoCon. For comparison, we add our polygon IoU loss to regress the 2D projected keypoints of 3D bounding boxes. We use the same training settings for both the baseline and modified model. The batch size is 8, and the total epoch number is 240. Following the original paper, We use an AdamW optimizer with a weight decay of 0.00001. We use a cyclic learning rate scheduler with an initial learning rate of 0.000225 which first gradually increases to 0.00225 with a step ratio of 0.4 and then gradually drops to $2.25e-8$. The experiments run on Ubuntu 20.04 and RTX A5000. Figure \ref{fig:visualizations} shows predicted outputs from the implemented model.

\subsection{MonoRCNN}

\subsubsection{Experiment Settings}
We modify the official code of MonoRCNN to incorporate our polygon IoU loss and use the original code to train the baseline. For comparison, we add PIoU loss to regress the 2D projected keypoints of 3D bounding boxes. We use the same training settings for both models. We train for 60000 iterations with a batch size of 8. The initial learning rate is 0.01 and is reduced by 0.1 after 30k, 40k, and 50k iterations. The experiments run on Ubuntu 20.04 and RTX A5000.

\section{Results} \label{sec:results}
Sections \ref{sec:results-cars} and \ref{sec:results-cyclists} present the results of each model on the KITTI cars and cyclists data, respectively, and Section \ref{sec:results-epochs} shows the performance changes for each model across various phases of training. Note that all AP scores are relatively low when compared with leading benchmark performance on the KITTI 3D detection dataset as a whole. This is because monocular 3D detection is relatively challenging when compared to 3D detection with sensor fusion as a whole  and viewpoint agnostic models additionally cannot incorporate scene information explicitly or implicitly into model structure or learning.

\subsection{Results on KITTI Cars} \label{sec:results-cars}
Table \ref{tab:cars-all} shows the final AP scores on KITTI 3D cars with different IoU thresholds. On RTM3D, PIoU loss has on average modestly better AP scores than the original model. Notably, $AP_{70}$ exclusively improves across the easy, moderate and hard subsets of the data, suggesting that PIoU offers the most benefit at higher IoU thresholds because L1 is often suitable for producing ``decent'' 3D detection results. Averaged across all difficulties, PIoU results in +0.18\% $AP_{70}$, +0.24\% $AP_{50}$, and +0.25\% $AP_{30}$. Proportional to baseline scores, $AP_{70}$. achieves the largest relative increase.

For the MonoCon model on KITTI cars, the proposed PIoU + L1 loss gives better results than the original model for nearly all different difficulty levels and at all IoU thresholds. (for the IoU threshold of 0.5, PIoU loss gives better results on the easy and moderate difficulty but not on the hard difficulty.) Again, the largest improvements from PIoU loss are seen at higher requisite IoU thresholds (+1.64\% $AP_{70}$, +0.64\% $AP_{50}$, and +0.64\% $AP_{30}$).

Lastly, the performance of the MonoRCNN model on KITTI cars is marginally worse when trained with PIoU + L1 loss versus L1 loss alone. However, there is still improvement at some IoU thresholds (-0.12\% $AP_{70}$, +0.42\% $AP_{50}$, and -1.42\% $AP_{30}$ averaged across difficulty levels), and the general trend across models holds that PIoU + L1 loss yields greater relative gain at stricter IoU thresholds.

Across all models, PIoU loss yields more improvements on easier difficulty levels. This trend is intuitive, as the ability of a loss function to improve performance is limited when features relevant to an object are not visible or are highly obscured; better model architectures can likely yield more performance improvement in these cases.

\begin{table}[hb]
\small
\begin{center}
\begin{tabular}{|l|lll|lll|}
\hline
\multirow{2}{*}{Model} & \multicolumn{3}{l|}{$AP_{50}$}                            & \multicolumn{3}{l|}{$AP_{25}$}                           \\ \cline{2-7} 
                                    & \multicolumn{1}{l|}{Easy} & \multicolumn{1}{l|}{Mod}  & Hard & \multicolumn{1}{l|}{Easy} & \multicolumn{1}{l|}{Mod}  & Hard \\ \hline \hline \hline
RTM3D                          & \multicolumn{1}{l|}{0.03} & \multicolumn{1}{l|}{0.03} & 0.03 & \multicolumn{1}{l|}{3.44} & \multicolumn{1}{l|}{2.00} & 1.56 \\ \hline
RTM3D* &
  \multicolumn{1}{l|}{\textbf{0.04}} &
  \multicolumn{1}{l|}{\textbf{0.04}} &
  \textbf{0.04} &
  \multicolumn{1}{l|}{\textbf{5.97}} &
  \multicolumn{1}{l|}{\textbf{3.17}} &
  \textbf{2.81} \\ \hline \hline \hline

MonoCon &
  \multicolumn{1}{l|}{4.25} &
  \multicolumn{1}{l|}{1.96} &
  1.90 &
  \multicolumn{1}{l|}{\textbf{19.20}} &
  \multicolumn{1}{l|}{\textbf{10.84}} &
  \textbf{10.22} \\ \hline
MonoCon* &
  \multicolumn{1}{l|}{\textbf{4.69}} &
  \multicolumn{1}{l|}{\textbf{2.63}} &
  \textbf{2.19} &
  \multicolumn{1}{l|}{18.27} &
  \multicolumn{1}{l|}{10.22} &
  9.53 \\ \hline \hline \hline

  MonoRCNN                       & \multicolumn{1}{l|}{3.05} & \multicolumn{1}{l|}{2.01} & 2.04 & \multicolumn{1}{l|}{15.98} & \multicolumn{1}{l|}{9.05} & 9.14 \\ \hline
MonoRCNN* &
  \multicolumn{1}{l|}{\textbf{4.31}} &
  \multicolumn{1}{l|}{\textbf{2.73}} &
  \textbf{2.55} &
  \multicolumn{1}{l|}{\textbf{19.05}} &
  \multicolumn{1}{l|}{\textbf{11.22}} &
  \textbf{11.28} \\ \hline
  
\end{tabular}
\end{center}
\caption{Evaluation results for RTM3D, MonoCon, and MonoRCNN on KITTI 3D Cyclist on the train/val split, evaluated by 3D $AP$ with IoU thresholds of 0.5 and 0.25. A * indicates models are trained with PIoU+L1 loss; other models are trained with L1 loss. Best result for each model at each threshold is shown in bold.}
\label{tab:cyclists-all}
\end{table}

\begin{table*}[htb]
\begin{center}
\begin{tabular}{|l|lll|lll|lll|}
\hline
\multirow{2}{*}{Model + Loss Types} &
  \multicolumn{3}{l|}{$AP_{50}$ @ 200 epochs} &
  \multicolumn{3}{l|}{$AP_{50}$ @ 160 epochs} &
  \multicolumn{3}{l|}{$AP_{50}$ @ 100 epochs} \\ \cline{2-10} 
 &
  \multicolumn{1}{l|}{Easy} &
  \multicolumn{1}{l|}{Mod} &
  Hard &
  \multicolumn{1}{l|}{Easy} &
  \multicolumn{1}{l|}{Mod} &
  Hard &
  \multicolumn{1}{l|}{Easy} &
  \multicolumn{1}{l|}{Mod} &
  Hard \\ \hline
RTM3D (L1) &
  \multicolumn{1}{l|}{24.13} &
  \multicolumn{1}{l|}{\textbf{18.62}} &
  \textbf{15.34} &
  \multicolumn{1}{l|}{24.29} &
  \multicolumn{1}{l|}{\textbf{18.74}} &
  \textbf{15.35} &
  \multicolumn{1}{l|}{17.66} &
  \multicolumn{1}{l|}{13.80} &
  11.15 \\ \hline
RTM3D (L1+PIoU) &
  \multicolumn{1}{l|}{\textbf{25.02}} &
  \multicolumn{1}{l|}{18.58} &
  15.25 &
  \multicolumn{1}{l|}{\textbf{24.91}} &
  \multicolumn{1}{l|}{18.59} &
  15.30 &
  \multicolumn{1}{l|}{\textbf{20.41}} &
  \multicolumn{1}{l|}{\textbf{15.11}} &
  \textbf{13.22} \\ \hline
Improvements &
  \multicolumn{1}{l|}{0.89} &
  \multicolumn{1}{l|}{-0.04} &
  -0.09 &
  \multicolumn{1}{l|}{0.62} &
  \multicolumn{1}{l|}{-0.15} &
  -0.05 &
  \multicolumn{1}{l|}{2.75} &
  \multicolumn{1}{l|}{1.31} &
  2.07 \\ \hline
\end{tabular}
\end{center}
\caption{Evaluation results for RTM3D on KITTI 3D Car on the train/val split, evaluated by $AP_{3D}$ with an IoU threshold of 0.5 at epoch 100, 160, and 200. Best result at each epoch is shown in bold.}
\label{tab:rtm3d_epochs}
\vspace{-0.01 in}
\end{table*}

\begin{table*}[htb]
\begin{center}
\begin{tabular}{|l|lll|lll|lll|}
\hline
\multirow{2}{*}{Model + Loss Types} &
  \multicolumn{3}{l|}{$AP_{70}$ @ 240 epochs} &
  \multicolumn{3}{l|}{$AP_{70}$ @ 160 epochs} &
  \multicolumn{3}{l|}{$AP_{70}$ @ 80 epochs} \\ \cline{2-10} 
 &
  \multicolumn{1}{l|}{Easy} &
  \multicolumn{1}{l|}{Mod} &
  Hard &
  \multicolumn{1}{l|}{Easy} &
  \multicolumn{1}{l|}{Mod} &
  Hard &
  \multicolumn{1}{l|}{Easy} &
  \multicolumn{1}{l|}{Mod} &
  Hard \\ \hline
MonoCon (L1) &
  \multicolumn{1}{l|}{22.27} &
  \multicolumn{1}{l|}{16.98} &
  14.69 &
  \multicolumn{1}{l|}{19.45} &
  \multicolumn{1}{l|}{14.69} &
  12.27 &
  \multicolumn{1}{l|}{3.47} &
  \multicolumn{1}{l|}{3.08} &
  2.45 \\ \hline
MonoCon (L1+PIoU) &
  \multicolumn{1}{l|}{\textbf{24.93}} &
  \multicolumn{1}{l|}{\textbf{18.45}} &
  \textbf{15.49} &
  \multicolumn{1}{l|}{\textbf{19.94}} &
  \multicolumn{1}{l|}{\textbf{15.35}} &
  \textbf{12.98} &
  \multicolumn{1}{l|}{\textbf{14.84}} &
  \multicolumn{1}{l|}{\textbf{10.84}} &
  \textbf{8.86} \\ \hline
Improvements &
  \multicolumn{1}{l|}{2.66} &
  \multicolumn{1}{l|}{1.47} &
  0.8 &
  \multicolumn{1}{l|}{0.49} &
  \multicolumn{1}{l|}{0.66} &
  0.71 &
  \multicolumn{1}{l|}{11.37} &
  \multicolumn{1}{l|}{7.76} &
  6.41 \\ \hline
  
\end{tabular}
\end{center}
\caption{Evaluation results for MonoCon on KITTI 3D Car on the train/val split, evaluated by $AP_{3D}$ with an IoU threshold of 0.7 at epoch 80, 160, and 240. Best result at each epoch is shown in bold.}
\label{monocon_epochs}
\end{table*}

\subsection{Results on KITTI Cyclists} \label{sec:results-cyclists}

Table \ref{tab:cyclists-all} compares the AP scores of incorporating PIoU loss versus L1 loss alone on KITTI 3D cyclists. For RTM3D, the accuracy is very low and therefore the change in prediction accuracy between the loss functions is negligible when IoU thresholds are 0.5., but PIoU+L1 loss significantly improves performance over L1 loss alone for an IoU threshold of 0.25 (+1.65\% $AP_{30}$ averaged across all difficulties.)

MonoCon performance is less notable on cyclists than on cars. Still, the performance of a model trained with PIoU+L1 loss is better than L1 alone at the more stringent IoU threshold of 0.5 (+0.56\% $AP_{50}$.)

For MonoRCNN, adding PIoU loss strictly improves cyclist detection performance. The AP scores in all IoU thresholds and difficulty levels increase by around 1-3\%. Here again, the largest improvements in prediction accuracy occur for the easy subsets of the data (+1.26\% $AP_{50}$ and +3.07\% $AP_{25}$).

\subsection{Model Performance at Different Epochs} \label{sec:results-epochs}
Lastly, we test the convergence speed of PIoU + L1 loss versus the baseline L1 loss. Table \ref{tab:rtm3d_epochs} shows the AP scores for RTM3D on KITTI cars at different epochs. When using PIoU loss, the AP scores early in training are significantly better than the baseline model (e.g. +2.75\% $AP_{50}$ on easy subset). (A similar trend is visible for $AP_{70}$ but this table is omitted for brevity). Table \ref{monocon_epochs} similarly shows PIoU+L1 loss results in faster model convergence than the L1 baseline at an IoU threshold of 0.7 for KITTI cars. After 80 epochs, PIoU results in at least +5\% $AP_{70}$ for each difficulty, with the largest performance improvement on the easy subset (+11.37\%). The results indicate strongly that PIoU+L1 loss converges faster than L1 loss alone. 

\section{Conclusions and Future Work} \label{conclusion}
In this work, we propose an efficient way to calculate IoU between convex polygons with irregular shapes. The batched implementation of the proposed PIoU loss in PyTorch is differentiable and can be used as the loss function for large object detection models. We show that PIoU loss can speed training convergence, both for simulated polygons and on the KITTI 3D detection dataset. We also show that using PIoU+L1 loss can increase the AP scores over L1 loss alone. Improvements vary when we incorporate PIoU loss into different 3D object detection models, with The CenterNet-based models benefitting more than the R-CNN-based model and a best result of +1.64\% $AP_{70}$ for MonoCon on KITTI cars. The most notable performance gains occur for highly visible vehicles when a strict IoU metric is required, meaning PIoU is especially helpful in transforming ``good'' predictions to ``great'' predictions.

This work tests PIoU loss on 3 different 3D object detection models for a benchmark dataset with relatively constrained (4-sided) polygons. In future work, we would like to incorporate the loss function into more difficult cases where rectangular bounding boxes are not sufficiently expressive and a detection formulation is preferred to a segmentation-based model formulation, providing an ``in-between'' for expressive detections of middling complexity. The results of PIoU in this work are quite promising, and we hope that future work can test the loss formulation on a larger sampling of methods and datasets, and in combination with other loss formulations.

\section*{Acknowledgements}
This work is supported by the National Science Foundation (NSF) under Grant No. 2135579, the NSF Graduate Research Fellowship Grant No. DGE-1937963 and the USDOT Dwight D. Eisenhower Fellowship program under Grant No. 693JJ32245006 (Gloudemans) and No. 693JJ322NF5201 (Wang). This material is based upon work supported by the U.S. Department of Energy’s Office of Energy Efficiency and Renewable Energy (EERE) award number CID DE-EE0008872. This material is based upon work supported by the CMAQ award number TN20210003. The views expressed herein do not necessarily represent the views of the U.S. Department of Energy, or the United States Government.

{\small
\bibliographystyle{ieee_fullname}
\bibliography{egbib}
}

\end{document}